\documentclass[letterpaper, 10 pt, conference]{ieeeconf}  

\IEEEoverridecommandlockouts               

\usepackage{amsmath}
\usepackage{graphicx}
\usepackage{amsfonts}
\usepackage{multirow}
\usepackage{xcolor}
\usepackage{booktabs}
\usepackage{float}
\usepackage{hyperref}
\usepackage{subcaption}
\usepackage{dblfloatfix} 
\usepackage{placeins}      

\hypersetup{hidelinks}
\graphicspath{{img/}}

\title{\LARGE \bf
Human-Interpretable Uncertainty Explanations \\ for Point Cloud Registration\thanks{$^{1}$University of T\"ubingen, T\"ubingen, Germany. $^{2}$Karlsruhe Institute of Technology (KIT), Karlsruhe, Germany. $^{3}$German Aerospace Center (DLR), We{\ss}ling, Germany. 
}
\thanks{This work was supported by EU Horizon project Inverse (grant agreement 101136067), MOTIE AiSac (Grant No. RS-2024-00441872), and DLR internal projects SKIAS 1.0 and 2.0. This work was also sponsored by the DFG SFB-1574 Circular Factory project and  the Ministry of Science, Research and Arts of the Federal State of Baden-Württemberg within the InnovationCampus Future Mobility.}
}

\author{Johannes A. Gaus$^{1,2}$, Loris Schneider$^{2}$, Yitian Shi$^{2}$,
Jongseok Lee$^{2,3}$, Rania Rayyes$^{2}$, Rudolph Triebel$^{2,3}$\\[2pt]{\tt\small johannes.gaus@uni-tuebingen.de}
}
\begin{document}

\maketitle
\bstctlcite{IEEEexample:BSTcontrol}
\thispagestyle{empty}
\pagestyle{empty}


\begin{abstract}

In this paper, we address the point cloud registration problem, where well-known methods like ICP fail under uncertainty arising from sensor noise, pose‐estimation errors, and partial overlap due to occlusion. We develop a novel approach, Gaussian Process Concept Attribution (GP-CA), which not only quantifies registration uncertainty but also explains it by attributing uncertainty to well-known sources of errors in registration problems. Our approach leverages active learning to discover new uncertainty sources in the wild by querying informative instances. We validate GP-CA on three publicly available datasets and in our real-world robot experiment. Extensive ablations substantiate our design choices. Our approach outperforms other state-of-the-art methods in terms of runtime, high sample-efficiency with active learning, and high accuracy. Our real-world experiment clearly demonstrates its applicability. Our video also demonstrates that GP-CA enables effective failure-recovery behaviors, yielding more robust robotic perception.

\end{abstract}
\section{Introduction}


Point cloud registration refers to the problem of estimating a relative transformation between two sets of 3D points~\cite{Besl1992}. This problem is essential in many robotic perception tasks, such as simultaneous localization and mapping (SLAM) \cite{zhang2014loam}, 3D reconstruction \cite{izadi2011kinectfusion}, and 6-DoF object pose estimation~\cite{sundermeyer2018implicit}, to name a few. A widely used approach for point cloud registration is Iterative Closest Point (ICP) \cite{rusinkiewicz2001}.  
However, in ICP, well-known causes
of uncertainty are 
sensor noise, poor initialization of the optimization process of the relative transformation between the two point clouds, and insufficient overlap between the two point clouds (e.g., due to occlusions), which can make ICP inaccurate or unreliable in practice  ~\cite{Maken2021}. Hence, several researchers attempt to improve the robustness of the registration process \cite{yang2020teaser}. 

Among others, several probabilistic approaches were proposed to quantify \textit{uncertainty in point cloud registration} \cite{Maken2021, de_maio_deep_2022, Landry2019, Bonnabel2016, Censi2007}. The underlying idea is that uncertainty estimates provide insight into the reliability of the obtained results, enabling the identification and rejection of unreliable registrations. 
Moreover, information about uncertainty is often used in many downstream tasks, including sensor fusion, state estimation, and 3D reconstruction~\cite{Maken2021}. To this end, a variety of tools have been developed to quantify uncertainty in point cloud registration, ranging from closed-form Gaussian solvers~\cite{Censi2007} to particle-based methods~\cite{Maken2021}.

\begin{figure}[!t]
  \centering
  \includegraphics[width=\columnwidth]{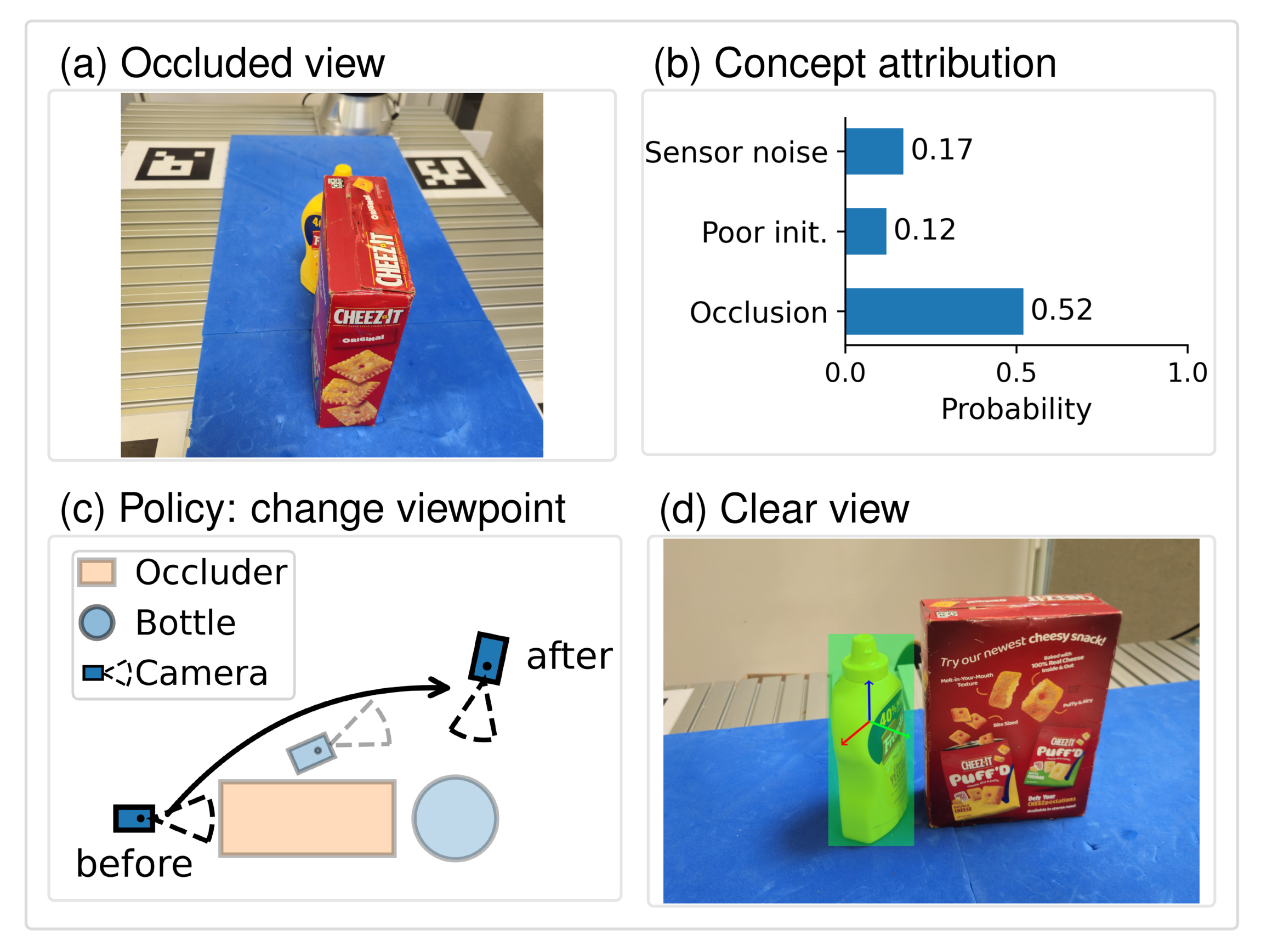} 
  \caption{Pose estimation via point-cloud registration. (a) The target object (mustard bottle) is occluded, so ICP-based 6-DoF pose estimation fails. (b) GP-CA attributes the failure primarily to occlusion. (c) The induced recovery policy is to change the viewpoint. (d) The new view reveals the bottle, ICP succeeds, and the bottle pose is successfully estimated.}
  \label{fig:pipeline1}
\end{figure}

While quantifying uncertainty helps to improve the registration, in practice, the magnitude of uncertainty is rarely sufficient in practical robotic perception tasks, as it does not reveal why registration failed or how to recover from it. Distinct failure causes — such as sensor noise, poor initialization, or occlusion \cite{Maken2021} — require different actions to recover from failure. 
In this work, we propose a novel approach, Gaussian Process Concept Attribution (GP-CA). A major difference to existing work is that, our approach not only quantifies uncertainty, but also explains uncertainty in a human-interpretable manner. Our GP-CA integrates active learning, learned point-cloud representations, and a Gaussian Process classifier to provide concept-level uncertainty attribution. The active learning allows fast adaptation to new uncertainty sources. 
In preliminary work \cite{Qin2024}, the explainability method SHAP was used to analyze the influence of different uncertainty sources on ICP-based registration. However, it cannot identify which uncertainty source is present in a registration and depends on manual control over uncertainty sources and heavy computation, limiting applicability and hindering its feasibility for robotic decision-making. 


GP-CA attributes uncertainty to semantically meaningful concepts that can be used later to enable robots to select targeted recovery actions. 
An example is depicted in Fig.~\ref{fig:pipeline1}, where the robot is tasked to perform a 6D pose estimation of objects using the ICP algorithm. Initially, an object of interest (the bottle) is occluded by another object, which causes the ICP algorithm to fail. Using our method, the occlusion is identified as the cause of failure, thereby enabling targeted recovery actions (e.g., changing the viewpoint). 
Hence, our main idea of explaining uncertainty in point cloud registration can improve the robustness of the perception tasks.

When registration fails or reports high uncertainty, GP-CA embeds the ICP-aligned source point cloud using a learned representation. Then, a multi-class Gaussian Process classifier (GPC) maps these representations to confidence scores over different concepts. In this way, we attribute the uncertainty to specific concepts, thereby identifying its source. The concepts are predefined by sets of examples. Still, a robot may encounter unknown sources of uncertainty during its operations. Therefore, GP-CA is equipped with an active learning mechanism \cite{narr2016stream, lee2025clever}, which can learn a new concept by querying the user for the label and choosing the most informative data to learn from. Across experiments on the LINEMOD \cite{Hinterstoisser2012}, the YCB \cite{Xiang2018}, and the Coffee Cup \cite{Lai2011} dataset and a custom real-world RGB-D dataset from our laboratory, we validate GP-CA and provide extensive ablation studies. We further propose how GP-CA can enable a more robust robotic perception by executing a set of failure recovery actions associated with uncertainty explanations. 

In summary, our main contributions are as follows.
\begin{itemize}
  \item We propose a novel approach, GP-CA, which enables human-interpretable uncertainty explanations in point cloud registration. To the best of our knowledge, this is the first work to show how explainability can advance point cloud registration for real-world applications
  \item We extend our approach with active learning to adapt and integrate new concepts.
  \item We validate the GP-CA design through an extensive ablation study
  \item We demonstrate clearly the runtime-efficiency, the high accuracy, and the sample-efficiency of our method, with SOTA baseline comparison across four datasets.
  \item We propose how GP-CA can be employed for recovery actions in a real-world robot setting.
\end{itemize}
The accompanying video illustrates the method and real robot experiments. 
\section{Related Work}
\label{sec:related_work}

Our method builds upon three main areas, namely uncertainty in point cloud registration, explainable AI, and active learning. We locate our contributions within these areas. 

\textbf{Uncertainty in Point Cloud Registration.} The ICP algorithm \cite{Besl1992,Chen1992} and its variants \cite{segal_generalized-icp_2009} are the gold standard in point cloud registration. In the past, many researchers have proposed to quantify uncertainty in ICP. Notably, closed-form covariance approximations \cite{Censi2007} and modeling of data-association effects \cite{Bonnabel2016} can be found. More recent data-driven methods \cite{Landry2019, de_maio_deep_2022} refine these estimates by learning covariances. Bayesian methods \cite{Maken2021} reformulate the optimization problem as probabilistic inference, where probability theory is used to make predictions based on available data. These directions quantify how much uncertainty is present rather than \textit{why} it arises. Our work complements these advances by explaining uncertainty with human-interpretable concepts.

\textbf{Explainable AI.} Explainable AI (XAI) generally refers to AI systems whose decision-making processes can be interpreted by humans \cite{ali2023explainable}. While models such as decision trees and rule-based systems are inherently explainable, XAI methods attempt to make black-box AI models such as neural networks more interpretable to humans. A widely known example are Saliency Maps \cite{ali2023explainable}, which visually explain deep learning models in image processing by highlighting which pixels of an input image contributed most to the output.

While Saliency Maps exploit the characteristics of neural networks, other XAI methods are designed to explain the predictions of any black-box model, which are also relevant to our problem. Among others, Shapley Additive Explanation (SHAP) and its variants \cite{lundberg_unified_2017} use game theory to explain the predictions of any model. SHAP approximates the importance of a specific input feature on the model output by comparing the outputs of many different combinations of features. Similarly, Sensitivity Analysis (SA) \cite{marino_methodology_2008} calculates the influence of input features via partial derivatives of a task metric (e.g., RMSE). Despite remarkable progress, these perturbation-based methods usually require heavy computation, which limits their practicality in online perception tasks. 
The most relevant methods for our use-case are concept-based methods, such as Testing with Concept Activation Vectors (TCAV) \cite{Kim2018}, which can provide explanations in terms of a model's reliance on concepts familiar to humans. To achieve this, TCAV trains a linear classifier on latent activations for positive and negative examples of a concept. Then, the direction in latent space that indicates that change in activation associated with this concept is computed from the decision boundary of the linear classifier. TCAV then measures output sensitivity to changes along this direction. This is done by calculating a score that indicates how important the concept is for the classification.

Unfortunately, directly employing TCAV for our problem of explaining uncertainty in ICP is non-trivial. We aim to explain why registration uncertainty is high, not directly which concept is most important for a classifier’s output. The first challenge is that ICP and its variants are geometric optimization algorithms, not neural networks. While they also provide a mapping from two point clouds to a transformation, we do not naturally have a latent space (or activations) on which TCAV relies to compute the concept scores. Second, the TCAV score lacks uncertainty estimates on its own. Beyond significance tests, TCAV does not provide calibrated probabilities over concepts, which we require to gate actions and defer decisions safely. Hence, while TCAV explains a model's prediction in human-friendly terms, it does not indicate how reliable these explanations are. Lastly, TCAV assumes a fixed set of predefined concepts and does not offer a mechanism to incorporate new concepts encountered online. Therefore, we propose GP-CA for human-interpretable, uncertainty-aware explanations in point cloud registration.

\textbf{Active Learning.} We apply active learning (AL) to improve our concept-based method. Active learning is an approach where the model chooses the most useful data to be labeled by an oracle (typically a human user). Specifically, we use Bayesian Active Learning by Disagreement (BALD) \cite{Houlsby2011}, but other active learning strategies can also be used. 
\section{Gaussian Process Concept Attribution}
\begin{figure}[t]
  \centering
  \includegraphics[width=\columnwidth]{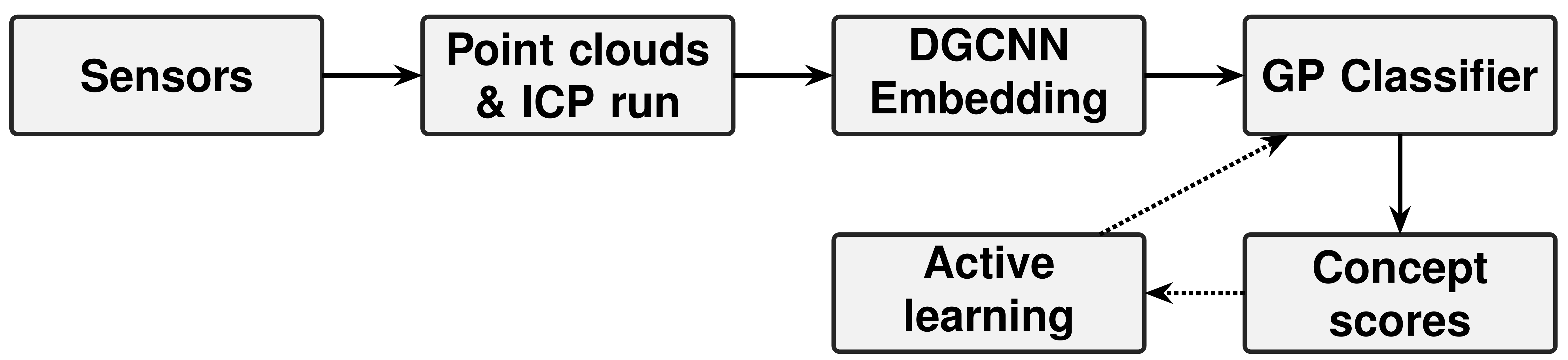}
  \caption{GP-CA Overview: ICP refines the pose, DGCNN encodes the point cloud into a latent vector. A Gaussian process classifier outputs concept probabilities with associated uncertainty. An active learning loop queries labels when uncertainty is high and updates the classifier.}
  \label{fig:sys_pipeline}
\end{figure}

Fig.~\ref{fig:sys_pipeline} provides an overview of our GP-CA approach. The input point cloud is aligned to a reference (e.g., ground truth or model frame) via ICP, which outputs a refined pose estimate. The point cloud is encoded by a Dynamic Graph CNN (DGCNN) \cite{Wang2019} into a latent vector, which a multi-class GPC maps to concept probabilities $\hat{s}_c$ with variances $v_c$ over noise, pose error, and partial overlap. 

\begin{figure}[t]
\centering
\begin{subfigure}[t]{0.32\columnwidth}
  \centering
\includegraphics[width=\linewidth,height=.9\linewidth,keepaspectratio,clip,trim=30 20 30 15]{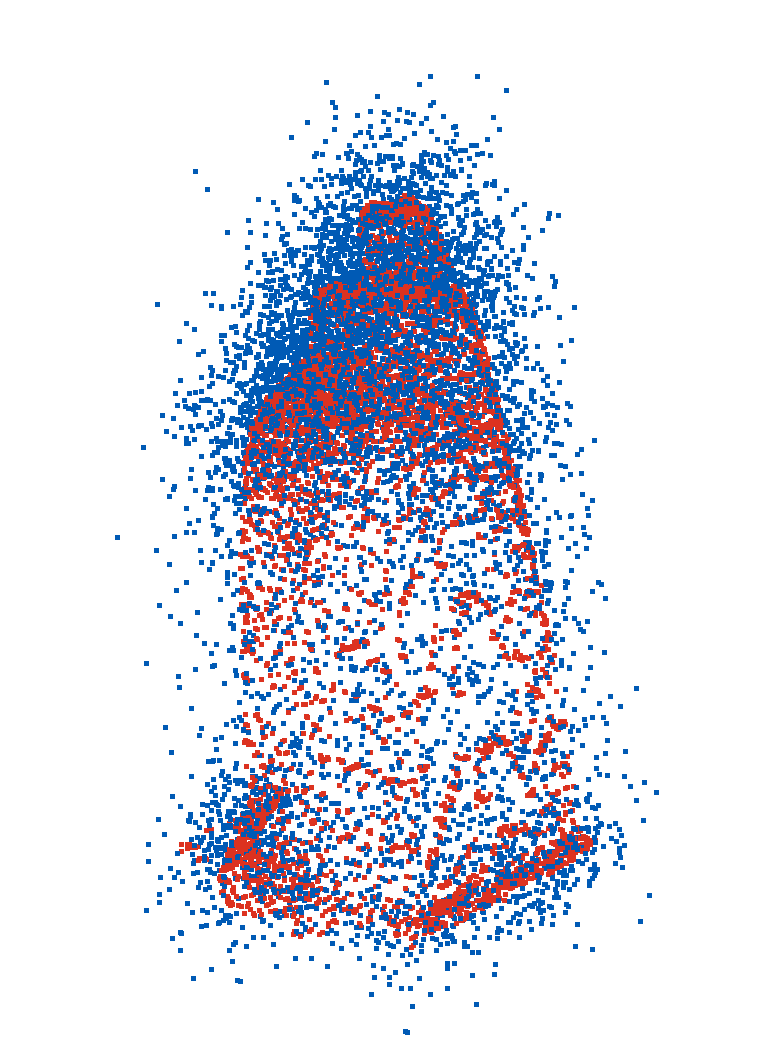}
  \caption{Noise}
\end{subfigure}\hfill
\begin{subfigure}[t]{0.32\columnwidth}
  \centering
\includegraphics[width=\linewidth,height=.9\linewidth,keepaspectratio,clip,trim=18 16 22 12]{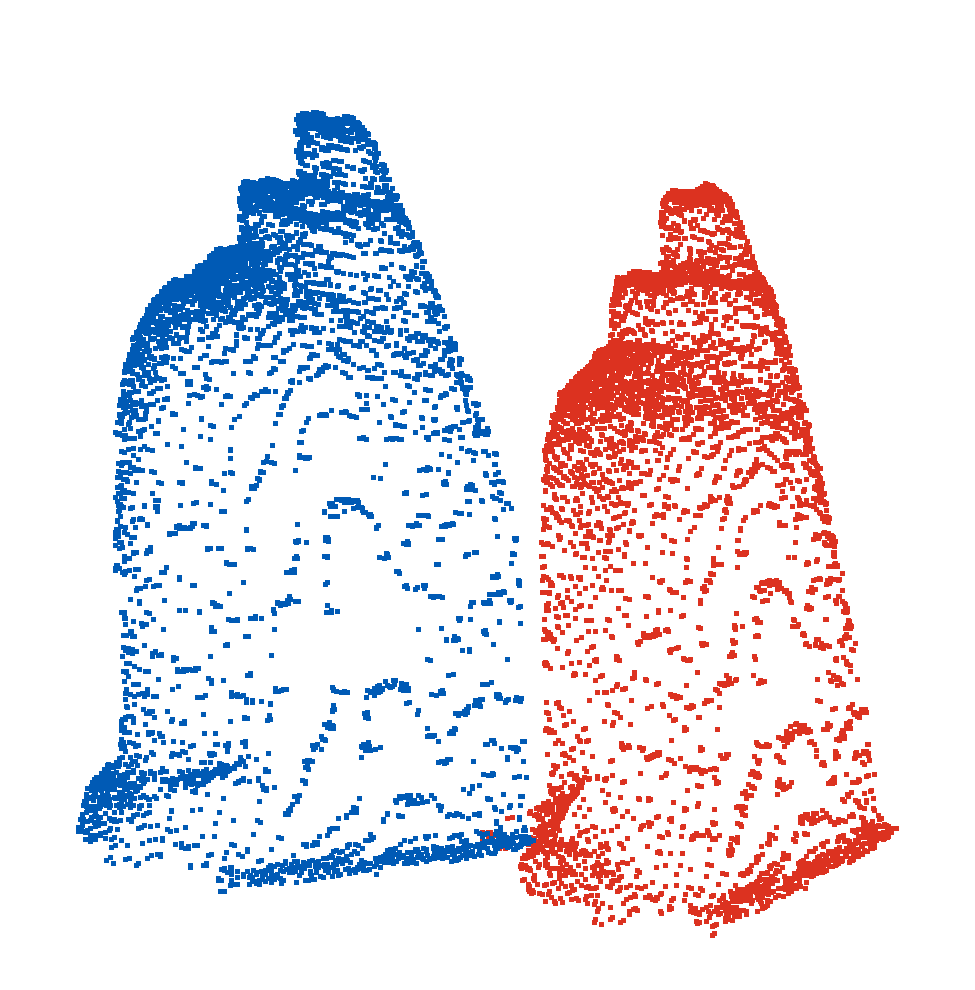}
  \caption{Pose error}
\end{subfigure}\hfill
\begin{subfigure}[t]{0.32\columnwidth}
  \centering
\includegraphics[width=\linewidth,height=.9\linewidth,keepaspectratio,clip,trim=28 22 28 20]{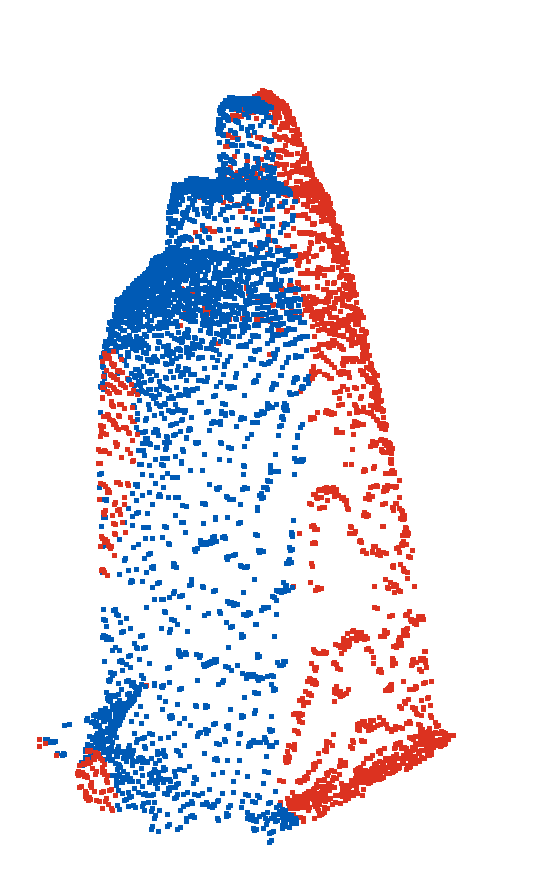}
  \caption{Overlap}
\end{subfigure}
\caption{Original point cloud (red) overlaid with a perturbed point cloud (blue) for each concept: noise (a), pose error (b) and partial overlap due to occlusion (c).}
\label{fig:concepts}
\end{figure}

Fig. \ref{fig:concepts} depicts three exemplary concepts, namely sensor noise, initialization error (or pose error in short), and lack of overlap between the two point clouds, which can make the ICP algorithm fail. Thus, when ICP returns high uncertainty, our framework analyzes the input point clouds and identifies the most likely concept that explains the failure. We use \emph{explainer} to denote the full GP-CA pipeline, from encoding the ICP-aligned source cloud with a DGCNN to mapping the resulting embedding to per-concept probabilities and associated variances.
We note that the encoder is kept fixed at inference time, and the classifier operates only on the embedding, which keeps the explainer gradient-free and light enough for real-time use. A simple active learning loop monitors uncertainty, requests a few targeted labels when needed, and updates the classifier so the explainer remains accurate as conditions change.

\subsection{Representation Learning in 3D}
We embed each registration instance with a DGCNN \cite{Wang2019} into a $d$-dimensional latent vector $\mathbf{h}=f_\theta(x)$. 
We pretrain the DGCNN by attaching a lightweight classification head and supervising it to predict the uncertainty concept from point clouds, then discard the head and freeze the backbone so its embedding specializes to concept-relevant geometry for the GPC. The concept classifier operates on a compact, high-level representation rather than raw points. Compared to PointNet~\cite{Qi2017} and PointNet++~\cite{Qi2017b}, which rely on global pooling and are less sensitive to local structure, the DGCNN better captures fine-grained geometry. 
In our ablations (Sec.~\ref{sec:ablations}), the DGCNN yields higher attribution accuracy and better calibration than PointNet~\cite{Qi2017} and other backbones under matched label and time budgets.

\begin{figure*}[!t] 
\centering 
\includegraphics[scale=0.055]{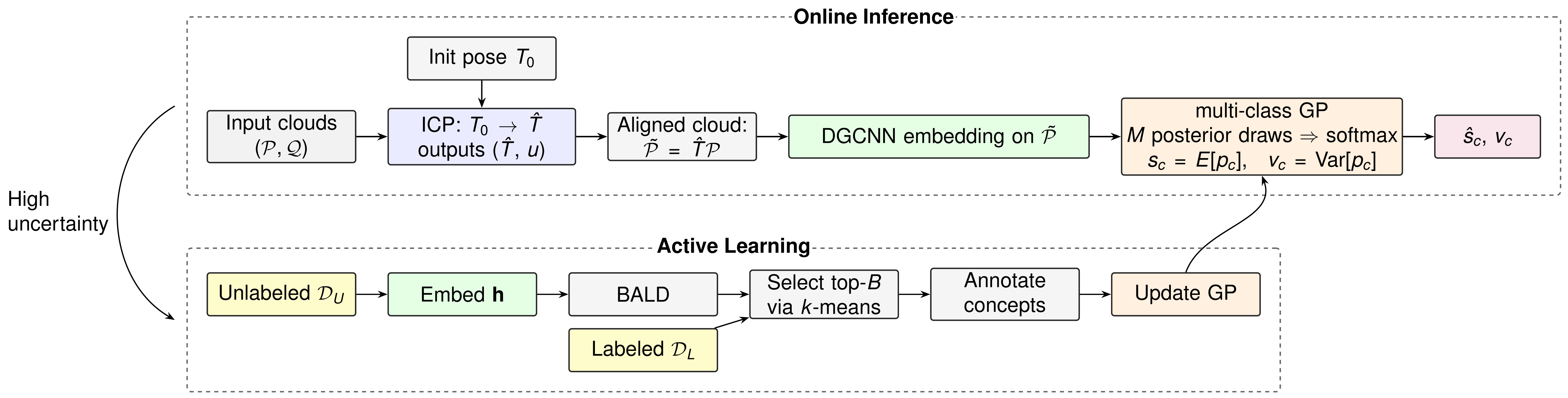} 
\caption{Pipeline with online inference (top) and active learning (bottom). \emph{Online Inference} (Sec.~\ref{sec:algo}): given input clouds $(\mathcal{P},\mathcal{Q})$ and an initial pose $T_0$, ICP returns $(\hat{T},u)$. We align the source $\tilde{\mathcal{P}}=\hat{T}\mathcal{P}$ and compute the DGCNN embedding $\mathbf{h}=f_\theta(\tilde{\mathcal{P}})$. A multi-class GPC (Sec.~\ref{sec:gpc}) draws $M$ posterior samples $\mathbf{f}^{(m)}(\mathbf{h})$; class probabilities use the softmax in Eq.~\eqref{eq:softmax}; calibrated scores $\hat{\mathbf{s}}$ and epistemic variances $\mathbf{v}$ follow Eqs.~\eqref{eq:mean}–\eqref{eq:var}; the predicted concept $\hat{c}$ uses Eq.~\eqref{eq:decision}. \emph{Active learning} (Sec.~\ref{sec:al}): we embed aligned clouds from an unlabeled pool, score uncertainty (e.g., BALD, Eq.~\eqref{eq:bald}), select a diverse top-$B$ via $k$-means, annotate concepts, and retrain the GPC.}
\label{fig:pipe}
\end{figure*}

\subsection{Gaussian Process Classification}
\label{sec:gpc}
We model concept attribution with a multi-class Gaussian Process classifier (GPC) \cite{Rasmussen2006}, which yields class probabilities with posterior (epistemic) uncertainty. Let $\mathcal{C}=\{\texttt{noise},\texttt{pose},\texttt{overlap}\}$ denote the concept set and let $\mathbf{h}\in\mathbb{R}^d$ be the DGCNN embedding. For each concept $c\in\mathcal{C}$ we place a GP prior over a latent logit function,
\begin{equation}
\{f_c\}_{c\in\mathcal{C}}, \qquad f_c \sim \mathrm{GP}\!\left(0,\,k_\phi(\cdot,\cdot)\right),
\label{eq:gp-prior}
\end{equation}
where $f_c(\mathbf{h})$ is a smooth latent logit that reflects the degree to which concept $c$ is represented in the instance $\mathbf{h}$, and $k_\phi$ is any positive-definite kernel with hyperparameters $\phi$ learned from data. Intuitively $f_c(\cdot)$ can be seen as a smooth “heatmap” over the embedding space, where higher values indicate stronger evidence for concept $c$. The kernel $k_\phi$ ensures that nearby embeddings produce similar outputs, with its hyperparameters controlling how smooth or sharp the variation is across the embedding space. The softmax likelihood maps logits to class probabilities:
\begin{equation}
p(y=c\mid \mathbf{f}(\mathbf{h})) \;=\;
\frac{\exp\!\big(f_c(\mathbf{h})\big)}{\sum_{k\in\mathcal{C}}\exp\!\big(f_k(\mathbf{h})\big)} ,
\label{eq:softmax}
\end{equation}
with $\mathbf{f}(\mathbf{h})=[f_c(\mathbf{h})]_{c\in\mathcal{C}}$. 
Only the differences between logits affect the final prediction.
Because the softmax likelihood in Eq.~\eqref{eq:softmax} is non-Gaussian (and thus non-conjugate to the GP prior), exact Bayesian inference is intractable: we cannot compute the posterior $p(\mathbf{f}\mid \mathcal{D})$ for a training dataset $\mathcal{D}$ in closed form, where $\mathcal{D}=\{(\mathbf{h}_i,y_i)\}_{i=1}^{N}$ denotes the labeled training set. We therefore introduce a tractable Gaussian variational posterior $q$ and write $p(\mathbf{f}\mid \mathcal{D}) \approx q(\mathbf{f})$.
Following \cite{hensman2015}, $q$ is parameterized by $N_Z$ shared inducing inputs $\mathbf{Z}=\{\mathbf{z}_i\}_{i=1}^{N_Z}$ and inducing values $\mathbf{u}_c=f_c(\mathbf{Z})$, which in turn define a predictive distribution $q(\mathbf{f}(\mathbf{h}))$ for any embedding $\mathbf{h}$.
These inducing inputs summarize the data, reducing GP training from cubic to linear in $N$ for fixed $N_Z$. We optimize the evidence lower bound (ELBO) with respect to the kernel parameters $\phi$, the inducing inputs $\mathbf{Z}$, and the variational parameters \cite{hensman2015}. 

For a test embedding $\mathbf{h}_*$, draw $\mathbf{f}^{(m)}(\mathbf{h}_*)\sim q(\mathbf{f}(\mathbf{h}_*))$ and compute class probabilities via Eq.~\eqref{eq:softmax}. Aggregating $M$ samples yields scores and variance by taking $p_c^{(m)}$, the predicted probability of concept $c$ for sample $m$, and calculating,
\begin{align}
\hat{s}_c &= \frac{1}{M}\sum_{m=1}^{M} p_c^{(m)}(\mathbf{h}_*), \label{eq:mean}\\
v_c &= \frac{1}{M}\sum_{m=1}^{M}\!\left(p_c^{(m)}(\mathbf{h}_*)-\hat{s}_c\right)^{\!2}. \label{eq:var}
\end{align}
Here, $\hat{\mathbf{s}}=[\hat{s}_c]_{c\in\mathcal{C}}$ is the concept attribution and $\mathbf{v}=[v_c]_{c\in\mathcal{C}}$ quantifies posterior uncertainty. The variance $v_c$ across samples is high when the model has seen only a few similar embeddings, and low when the data is dense. In practice, we select the concept with the largest probability,
\begin{equation}
\hat{c}=\arg\max_{c\in\mathcal{C}} \hat{s}_c\, ,
\label{eq:decision}
\end{equation}
and can defer when uncertainty is high.

\subsection{Active Learning}
\label{sec:al}
Concept-based explanations must remain valid as the environment changes. However, new sources of uncertainty can degrade a static classifier over time. To maintain accuracy, we employ a lightweight active learning loop that adapts the GPC based on targeted feedback, as illustrated in Fig.~\ref{fig:pipe} (\emph{Active Learning}).
When the GPC expresses high uncertainty on new data, we trigger an active learning loop to improve robustness. The algorithm queries the user to label new informative examples from an unlabeled dataset $\mathcal{D}_U$, which are added to a labeled dataset $\mathcal{D}_l$. The GPC is then retrained using both previously labeled data and the newly annotated points. Since GPC training is lightweight, retraining remains fast as the dataset grows. In this way, the model gradually adapts to new concepts and improves prediction accuracy.

To decide which points to annotate, we use the Bayesian Active Learning by Disagreement (BALD) criterion \cite{Houlsby2011}. A high BALD score indicates disagreement among posterior samples, highlighting points whose labeling most reduces uncertainty. We compute the BALD score \cite{Houlsby2011} for each unlabeled embedding $\mathbf{h}$ as:
\begin{equation}
\mathrm{BALD}(\mathbf{h}) \;=\; H\!\big[\hat{\mathbf{s}}(\mathbf{h})\big]
\;-\; \frac{1}{M}\sum_{m=1}^{M} H\!\big[\mathbf{p}^{(m)}(\mathbf{h})\big].
\label{eq:bald}
\end{equation}
where $H[\cdot]$ denotes entropy, $\mathbf{p}^{(m)}(\mathbf{h})$ are class probabilities from posterior sample $m$, and $\hat{\mathbf{s}}(\mathbf{h})$ is the predictive mean from Eq.~\eqref{eq:mean}. The BALD score is high when posterior samples disagree with the mean prediction, indicating high epistemic uncertainty and expected information gain. However, BALD alone often selects very similar datapoints, as high uncertainty tends to cluster in specific regions of the embedding space. This leads to redundant queries that waste labeling effort. 
To reduce redundancy, we apply a two-step \emph{score-then-diversify} strategy. Let \( B \) denote the number of points that need to be labeled, i.e., annotation budget per round, and \( K>1 \) a small expansion factor.  \\
\emph{1) Score:} Compute BALD scores (Eq.~\eqref{eq:bald}) for all unlabeled embeddings and select the top \( B \cdot K \) embeddings.\\
\emph{2) Diversify:} Cluster these \( B \cdot K \) candidates into \( B \) clusters with \( K \) datapoints per cluster using $k$-means in embedding space.\\
Then we select the point closest to each cluster centroid. The selected \( B \) points, as representatives for each cluster,  are then labeled, added to the labeled set \( \mathcal{D}_l \), and the GPC is retrained. This ensures both informativeness (via BALD) and diversity (via clustering), avoiding redundant queries. In our experiments (Sec.~\ref{sec:ablations}), this strategy consistently outperformed random selection, which is a typical baseline for active learning.

\subsection{Inference}
\label{sec:algo}
Fig.~\ref{fig:pipe} (\emph{Online Inference}) shows the inference procedure. We receive a source cloud $\mathcal{P}$, a target cloud $\mathcal{Q}$, and an initial pose $T_0$. ICP refines $T_0$ to $\hat{T}$ and returns a registration uncertainty $u$. We apply $\hat{T}$ to the source to obtain the aligned cloud $\tilde{\mathcal{P}}=\hat{T}\mathcal{P}$ and compute the DGCNN embedding from this final cloud, $\mathbf{h}=f_\theta(\tilde{\mathcal{P}})$. From $\mathbf{h}$, we obtain concept scores and their variances, $\hat{\mathbf{s}}$ and $\mathbf{v}$, as described in Sec.~\ref{sec:gpc}. 
Updates happen occasionally when high uncertainties accumulate. We score an unlabeled pool with the BALD acquisition (cf.~Sec.~\ref{sec:al}), select a small and diverse batch by clustering high-scoring candidates, collect concept labels, and update the GPC. The encoder can remain frozen or be lightly fine-tuned if needed and only the GPC is updated during active learning.

\section{Experimental Results}
\label{sec:experiments}
We present our experimental results in five steps. First, we provide an overview of the datasets used and describe the process of synthesizing perturbed point clouds in Sec.~\ref{sec:datasets}. Then, in Sec.~\ref{sec:RQ1}, we compare the single-concept attribution accuracy of GP-CA with XAI methods introduced in Sec. \ref{sec:related_work}. Next, we test the label efficiency of active learning with BALD against random sampling in Sec. \ref{sec:rq2}, as it is a common baseline in active learning. In Sec. \ref{sec:ablations}, we substantiate design choices through ablation studies. Finally, we validate the transfer to real-world scenes and examine the runtime of each part of GP-CA in Sec. \ref{sec:RQ4}.


\subsection{Datasets and Controlled Perturbations}
\label{sec:datasets}
To evaluate our method's ability to attribute uncertainty to semantic causes, we construct synthetic datasets from three real-world 3D object collections: a Coffee Cup set \cite{Lai2011}, LINEMOD \cite{Hinterstoisser2012}, YCB \cite{Xiang2018}, and our own real-world data consisting of point clouds extracted from RGB-D images. For each object, we generate point cloud pairs $(\mathcal{P}, \mathcal{Q})$, where $\mathcal{Q}$ is derived from $\mathcal{P}$ by applying a single controlled perturbation to isolate specific sources of registration uncertainty:\\
-- \textbf{Noise}: Gaussian jitter added to all points in $\mathcal{P}$ to simulate sensor-level disturbances.\\
-- \textbf{Pose}: Random rigid-body transformations applied to $\mathcal{P}$, introducing variation in rotation and translation.\\
-- \textbf{Overlap}: Random cropping of points in $\mathcal{P}$ to emulate partial visibility or occlusion.\\
Fig. \ref{fig:concepts} shows examples of point cloud pairs from our dataset for each type of perturbation. Each sample is labeled with the single concept used to generate the perturbation, enabling controlled supervision for attribution. We generate 800 training and 200 test samples per dataset, balanced across the three concept types. This setup allows precise evaluation of concept attribution under isolated semantics.

\subsection{Concept Attribution Accuracy on Datasets}
\label{sec:RQ1}
To contextualize GP-CA, we compare it against established XAI methods (see Sec.~\ref{sec:related_work}). Following \cite{Qin2024}, we measure the uncertainty of the ICP by computing multiple pose estimates and calculating the Kullback-Leibler divergence. The three sources of uncertainty are modeled as input features whose influence on the measured uncertainty is estimated by SHAP \cite{lundberg_unified_2017} and SA \cite{marino_methodology_2008}. 

When comparing the accuracy of the uncertainty attribution when only a single concept is present in each input sample, SHAP and SA perform poorly compared to GP-CA as reported in Tab.~\ref{tab:single_baselines}. Each experiment is repeated five times per dataset and we report the mean accuracy per dataset.

To make the baselines as fast as possible we limited SHAP coalition sampling and feature subsets and used low-resolution perturbation sweeps for SA. Even with these speed-oriented settings, SHAP still required 85\,s to 125\,s per sample and SA 20\,s to 25\,s per sample, compared to 2\,s to 4\,s per sample for GP-CA.

\begin{table}[h]
\centering
\caption{Comparison of mean accuracy of single-concept attribution over five experiments on each dataset (\%).}
\label{tab:single_baselines}
\small
\setlength{\tabcolsep}{6pt}
\begin{tabular}{lccc}
\toprule
\textbf{Dataset} & \textbf{SHAP} & \textbf{SA} & \textbf{GP-CA} \\
\midrule
Coffee Cup & 33.3 & 19.2 & \textbf{97.7} \\
LINEMOD    & 38.3 & 17.0 & \textbf{94.0} \\
YCB        & 31.4 &  7.9 & \textbf{99.6} \\
\bottomrule
\end{tabular}
\end{table}

We present the results for GP-CA in more detail and report the mean per-concept accuracy over the five experiments on each dataset in Tab.~\ref{tab:single}. It shows that GP-CA achieves near-perfect attribution across datasets, indicating that noise, pose, and overlap can be clearly separated.

\begin{table}[t]
\centering
\caption{Mean per-concept accuracy of GP-CA over five experiments on each dataset (\%).}
\label{tab:single}
\small
\setlength{\tabcolsep}{5pt}
\begin{tabular}{lcccc}
\toprule
\textbf{Dataset} & \textbf{Noise} & \textbf{Pose} & \textbf{Overlap} \\
\midrule
Coffee Cup & \textbf{100.0} & 99.1 & 93.0 \\
LINEMOD    &  94.0 & 89.0 & \textbf{100.0}\\
YCB        &  99.0 & \textbf{100.0} & \textbf{100.0} \\
\bottomrule
\end{tabular}
\end{table}

These findings illustrate that GP-CA can provide concept-aware explanations with practical relevance. As presented in Tab.~\ref{tab:single}, GP-CA could capture clearly which concept causes the high uncertainty. Across three publicly available datasets, we show that GP-CA can achieve significantly higher accuracy compared to applying SHAP and SA. 

\subsection{Active Learning via BALD}
\label{sec:rq2}
We study sample efficiency when the causes of uncertainty are not fully known in advance. In practice, new sensors, scenes, or deployment conditions can introduce shifts or new failure modes, so the explainer must learn efficiently and adapt as concepts evolve. To replicate this case, we use a GPC trained only on examples of noise and pose perturbations but validate it on a dataset containing all three concept classes. When the framework encounters regions with high posterior variance, it queries labels. We compare the selection of embeddings for labeling with BALD and $k$-means diversification with randomly selecting embeddings. To evaluate efficiency, we count the number of labels required to reach 90\% concept attribution accuracy on the validation dataset for each selection method.

\begin{table}[h]
\centering
\caption{Label efficiency and convergence. Columns show labels needed to reach 90\% accuracy, final validation accuracy after AL, and epochs to converge.}
\label{tab:al_labels}
\scriptsize
\setlength{\tabcolsep}{4pt}
\begin{tabular}{lcccc}
\toprule
\textbf{Dataset} & \textbf{\#Labels BALD} & \textbf{\#Labels Random }&\textbf{Acc. (\%)} & \textbf{Epochs} \\
\midrule
Coffee Cup & 12 & 350--550 & 100.0 & $\sim$60--80 \\
LINEMOD    & 20 & 350--550 & 95.6  & $\sim$100 \\
YCB        & 14 & 350--550 & 100.0 & $\sim$60--80 \\
\bottomrule
\end{tabular}
\end{table}

Tab.~\ref{tab:al_labels} summarizes label efficiency and convergence for active learning. The first two columns report the number of labels required to reach 90\% attribution accuracy for BALD versus random sampling. All values are averaged over five runs, and label counts are rounded to the nearest integer. BALD needs only 12/20/14 new labeled datapoints on Coffee Cup/LINEMOD/YCB, respectively to add a new concept to the GPC, while random sampling requires 350--550 new labeled datapoints across datasets. The third column lists the final validation accuracy after active learning, and the last column shows the approximate training epochs to reach the performance of 90\% accuracy. Overall, BALD attains 100.0/95.6/100.0\% on Coffee Cup/LINEMOD/YCB, respectively, with convergence in roughly 60--100 epochs, confirming fast, label-efficient learning.
In our active-learning experiments, GP-CA improves attribution accuracy by querying labels only where the GPC is uncertain. The GPC’s per-concept uncertainties drive this targeted retraining that standard explainers generally lack. As a result, GP-CA adapts to previously unseen conditions and expands its concept vocabulary over time by acquiring and integrating new sources of uncertainty.

\subsection{Ablation Studies}
\label{sec:ablations}
We evaluate three design choices. (i) The GPC’s kernel and optimization schedule; (ii) the learner; (iii) the feature extraction.

\textbf{Kernel.} We evaluate three different kernel choices for the GPC: a RBF kernel, a Matérn-3/2 kernel and a polynomial kernel. On the YCB dataset, the RBF kernel with a learning rate of 0.01 reaches 99.0\% validation accuracy after 100 iterations; the best YCB accuracy is 99.3\% (tied for Matérn–3/2 and Poly, see Tab.~\ref{tab:abl_kern}). On LINEMOD, Matérn–3/2 with the same learning rate and 200 iterations reaches an accuracy of 97.8\%. Very small learning rates (0.001) underfit on both datasets.
Our experiments show that in our pipeline, GPC can be equipped with different kernel functions with only minor influence on accuracy.
\vspace{-4pt}
\begin{table}[h]
\centering
\caption{Kernel and schedule. Validation accuracy (\%). Y/L = YCB/LINEMOD.}
\label{tab:abl_kern}
\small
\setlength{\tabcolsep}{3.2pt}
\renewcommand{\arraystretch}{0.95}
\begin{tabular}{lcc}
\toprule
\textbf{Setting} & \textbf{Iter.} &\textbf{ Y/L (\%)} \\
\midrule
RBF, lr=.01                      & 100 & 99.0 / 93.9 \\
Matérn $\nu{=}1.5$, lr=.01       & 200 & \textbf{99.3} / \textbf{97.8} \\
Poly. $p{=}2$, lr=.01             & 200 & \textbf{99.3} / 94.4 \\
RBF, lr=.001                     & 200 & 69.3 / 50.6 \\
Poly., $p{=}2$, lr=.001           & 200 & 88.0 / 53.9 \\
\bottomrule
\end{tabular}
\end{table}
\vspace{-4pt}

\textbf{Learner.} We compare using a multi-class GPC with using a Random Forest (RF) and a Support Vector Machine (SVM). Following \cite{Grimmett2016}, we report three diagnostics.

The Spearman rank correlation $\rho$ is computed between the per-sample uncertainty and the distance of each sample to the training distribution. A high $\rho$ indicates that the model assigns higher uncertainty to out-of-distribution points. We check whether samples far from the training distribution are ranked as most uncertain. The AUC is the Area Under the Coverage-Risk Curve for selective prediction. Lower values indicate better rejection of uncertain predictions while keeping coverage. As we raise the confidence threshold and keep fewer predictions, a good model’s error should drop quickly and a lower AUC reflects this. ECE (Expected Calibration Error) measures how well predicted probabilities match observed accuracy across confidence bins. It is computed by grouping predictions by confidence and comparing each bin’s average confidence to its empirical accuracy; the average gap is the ECE. Smaller ECE indicates better calibration.

The multi-class GPC matches or exceeds RF and SVM in accuracy and produces uncertainty that tracks distributional shift and supports active selection. On the YCB dataset, it achieves a Spearman rank correlation of $\rho=0.592$ and a coverage–risk area of $\mathrm{AUC}=0.0078$. RF is close in accuracy, but yields weaker uncertainty, with $\rho=0.132$ and a larger AUC. SVM shows lower ECE, yet its Spearman rank correlation is near zero or negative ($\rho=-0.011$), which is not useful for acquisition. We adopt the full multi-class GPC as the default learner. All baselines used the \texttt{scikit-learn} \cite{scikit-learn} defaults: RF with \texttt{n\_estimators=50} and \texttt{random\_state=42}, and a linear SVM with \texttt{random\_state=42}. The results are in Tab.~\ref{tab:abl_models}.

In summary, while RF and SVM are competitive in accuracy, their uncertainty estimates are less reliable. In contrast, GPC not only maintains high accuracy but also provides well-calibrated uncertainty estimates that robustly track distributional shift and enable effective active acquisition.

\begin{table}[t]
\centering
\caption{Validation accuracy (\%) and uncertainty diagnostics (YCB). $\uparrow$ indicates higher is better; $\downarrow$ indicates lower is better.}
\label{tab:abl_models}
\small
\setlength{\tabcolsep}{3pt}
\renewcommand{\arraystretch}{0.92}
\begin{tabular}{lccc c}
\toprule
& \multicolumn{3}{c}{\textbf{Val Acc (\%)}} & \textbf{$\rho\uparrow,\ \text{AUC}\downarrow,\ \text{ECE}\downarrow$} \\
\cmidrule(lr){2-4}
\textbf{Model} & CC & LM & YCB &   \\
\midrule
GPC        & 100.0 & \textbf{97.8} & \textbf{99.3} & \textbf{0.592}, \textbf{0.0078}, 0.223 \\
RF         & 100.0 & 96.7 & 98.0 & 0.132, 0.1145, 0.218 \\
Linear SVM & 100.0 & 86.1 & 98.0 & -0.011, 0.0168, \textbf{0.104} \\
\bottomrule
\end{tabular}
\end{table}

\textbf{Feature Extraction Backbone.} We compare the learned feature embeddings from the DGCNN to classic analytical features based on statistics. Those features include terms such as RMSE, normal consistency, planarity, linearity, and point-count ratios. In our experiments, the analytical features cap accuracy near 66\% on the YCB dataset, while the DGCNN embedding reaches about 99\% under the same GPC settings and label budget. The learned embeddings yield features with higher separability and better calibration.



We evaluated the accuracy when deploying a DGCNN against other backbones (i.e. PointNet \cite{Qi2017}, PointNet++ \cite{Qi2017b} and PointNeXt \cite{Qian2022}). We also compared the influence of compressing the features using dimensionality reduction methods, namely Principal Component Analysis (PCA) \cite{Jolliffe2002}, Uniform Manifold Approximation and Projection (UMAP) \cite{mcinnes_umap_2018} and a trained Auto Encoder (AE) \cite{Hinton2006}. For each combination of backbone and reduction method, five structurally identical multi-class GPCs with RBF kernels were trained. The GPCs were evaluated on 300 samples from our real-world dataset. Tab.~\ref{tab:abl_feat_rw} shows the mean accuracy and standard deviation over the five evaluations for each backbone and reduction method. We find that DGCNN yields a higher accuracy than other backbones, with different dimensionality reduction methods as well as without. DGCNN achieves the highest accuracy (85.2\%) without dimensionality reduction.  Applying PCA or a lightweight autoencoder produces only minor drops (84.87\% and 83.67\%, respectively), while PointNet, PointNet++, and PointNeXt trail by a considerable margin under identical GPC settings.
The results show that using DGCNN features without any dimensionality reduction yields a higher accuracy compared to analytical features or features extracted from PointNet, PointNet++ or PointNeXt.

\begin{table}[h]
\centering
\caption{Comparison of different backbone architectures and dimensionality reduction methods.}
\label{tab:abl_feat_rw}
\small
\setlength{\tabcolsep}{3pt}
\renewcommand{\arraystretch}{0.95}
\begin{tabular}{llcc}
\toprule
\textbf{Backbone} & \textbf{Reduct. Method} & \textbf{Mean Acc. (\%)} & \textbf{Std. Dev. (\%)}\\
\midrule
DGCNN & No Reduction & \textbf{85.2} & 1.98\\
 & PCA & 84.9 & 1.93\\
 & UMAP & 83.2 & 1.87\\
 & AE & 83.7 & \textbf{1.72}\\
\midrule
PointNet & No Reduction & 69.7 & \textbf{4.16}\\
 & PCA & \textbf{79.3} & 4.18\\
 & UMAP & 61.5 & 5.46\\
 & AE & 50.7 & 12.53\\
\midrule
PointNet++ & No Reduction & 59.8 & 5.43\\
 & PCA & \textbf{71.4} & \textbf{1.82}\\
 & UMAP & 68.1 & 4.12\\
 & AE & 43.2 & 5.56\\
\midrule
PointNeXt & No Reduction & 69.6 & 4.94\\
 & PCA & \textbf{83.5} & \textbf{2.84}\\
 & UMAP & 64.5 & 5.64\\
 & AE & 42.4 & 4.21\\
\bottomrule
\end{tabular}
\end{table}

Based on our findings, we adopt a multi-class GPC with a RBF kernel on DGCNN features without any dimensionality reduction. We train with 100 epochs by default and rely on BALD for label-efficient updates during active learning.


\subsection{Real-World Validation}
\label{sec:RQ4}
We recorded a large in–lab RGB-D corpus with varied objects, backgrounds, and viewpoints. Each sequence provides depth maps and camera poses to extract point clouds for ICP. The pipeline mirrors deployment: segmentation and initial 6-DoF pose from FoundationPose \cite{Wen2024}, ICP refinement, DGCNN feature extraction, and GP-CA concept attribution.

The best-performing configuration is a DGCNN backbone without dimensionality reduction, achieving 85.2\% overall accuracy on the real-world set (Tab.~\ref{tab:abl_feat_rw}). It yields 99.4\% accuracy on noise perturbations, 78.8\% accuracy on pose perturbations, and 76.9\% accuracy on overlap perturbations. A major source of misclassification is pairwise confusion between pose and overlap perturbations. Specifically, 20.6\% of true pose perturbations were misclassified as overlap (false negatives for pose), while 23.0\% of true overlap perturbations were misclassified as pose (false negatives for overlap).
Misclassifications typically occur in scenes where concepts co-occur (e.g., partial overlap plus a small pose error), reflecting the fluid boundaries between causes.

\begin{figure}[t]
\centering
\newcommand{\panelw}{0.48\columnwidth}
\newcommand{\panelh}{0.35\columnwidth}
\captionsetup[sub]{position=top, font=normalsize, labelfont=bf, skip=1pt}
\setlength{\tabcolsep}{1pt}
\setlength{\fboxsep}{0pt}
\setlength{\fboxrule}{0.001pt}
\begin{tabular}{@{} c c @{}} 
  \subcaptionbox{Input \& ROI\label{fig:robot:a}}[\panelw]{%
    \fbox{\includegraphics[width=\panelw,height=\panelh]{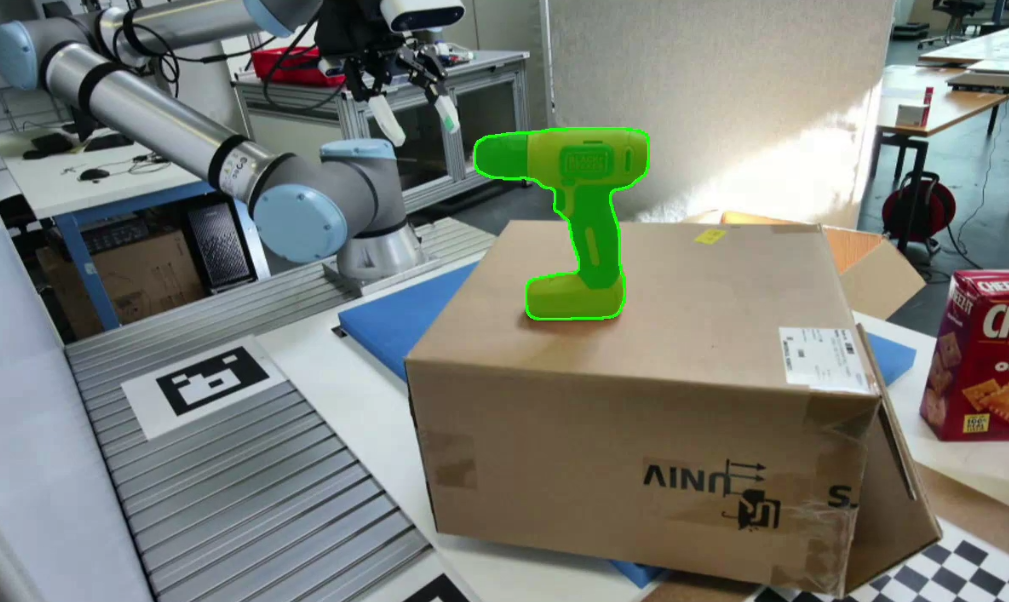}}%
  } &
  \subcaptionbox{Initial pose\label{fig:robot:b}}[\panelw]{%
    \fbox{\includegraphics[width=\panelw,height=\panelh]{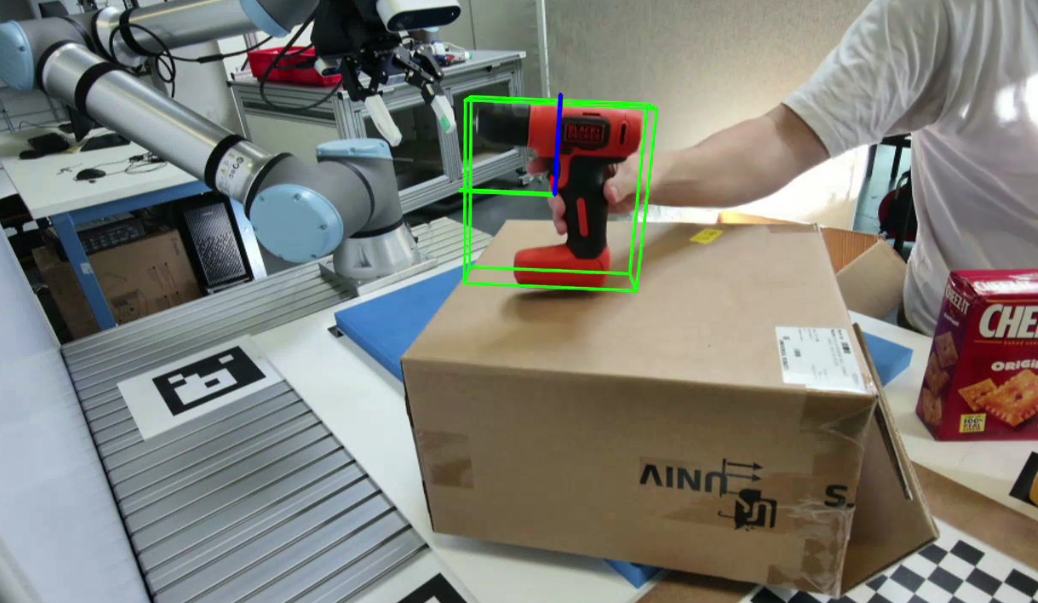}}%
  } \\
\noalign{\vspace{3pt}}
  \subcaptionbox{ICP alignment\label{fig:robot:c}}[\panelw]{%
    \fbox{\includegraphics[width=\panelw,height=\panelh]{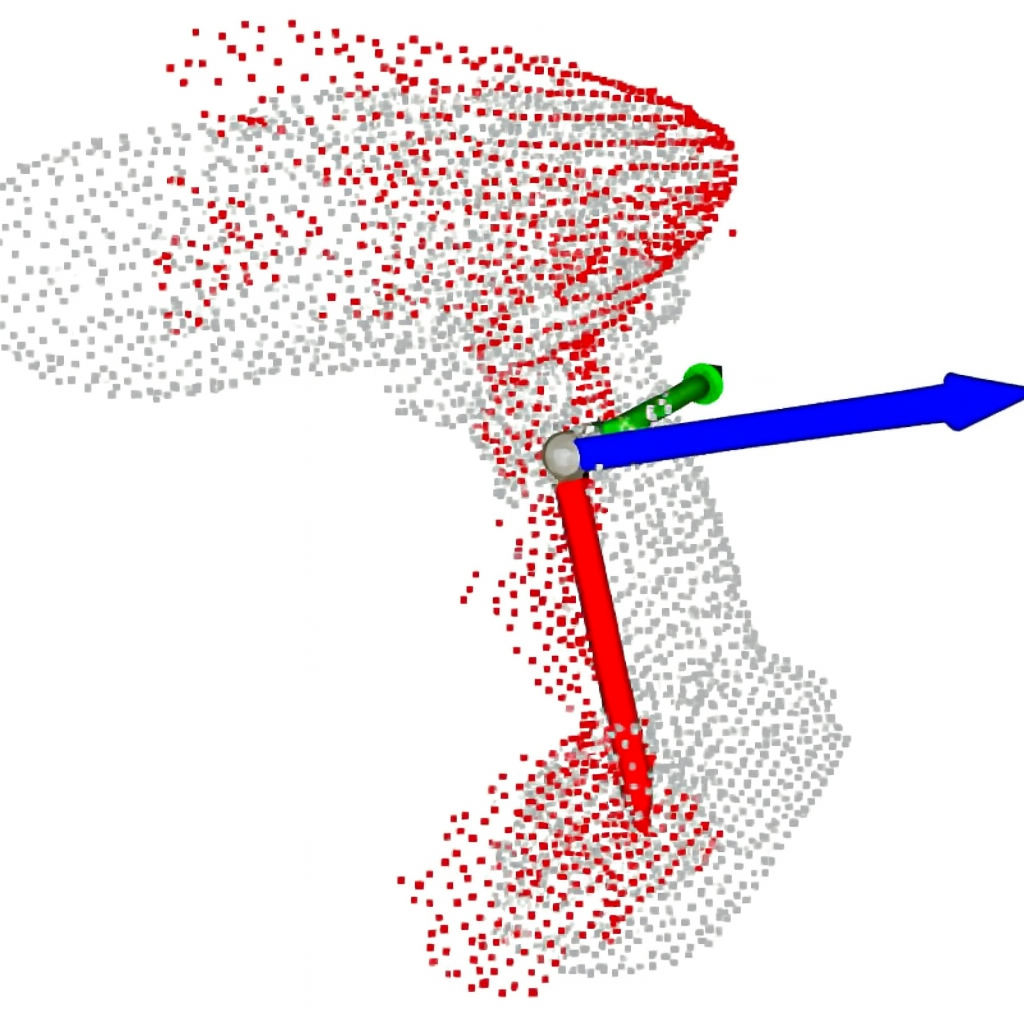}}%
  } &
  \subcaptionbox{GP-CA attribution\label{fig:robot:d}}[\panelw]{%
    \fbox{\includegraphics[width=\panelw,height=\panelh]{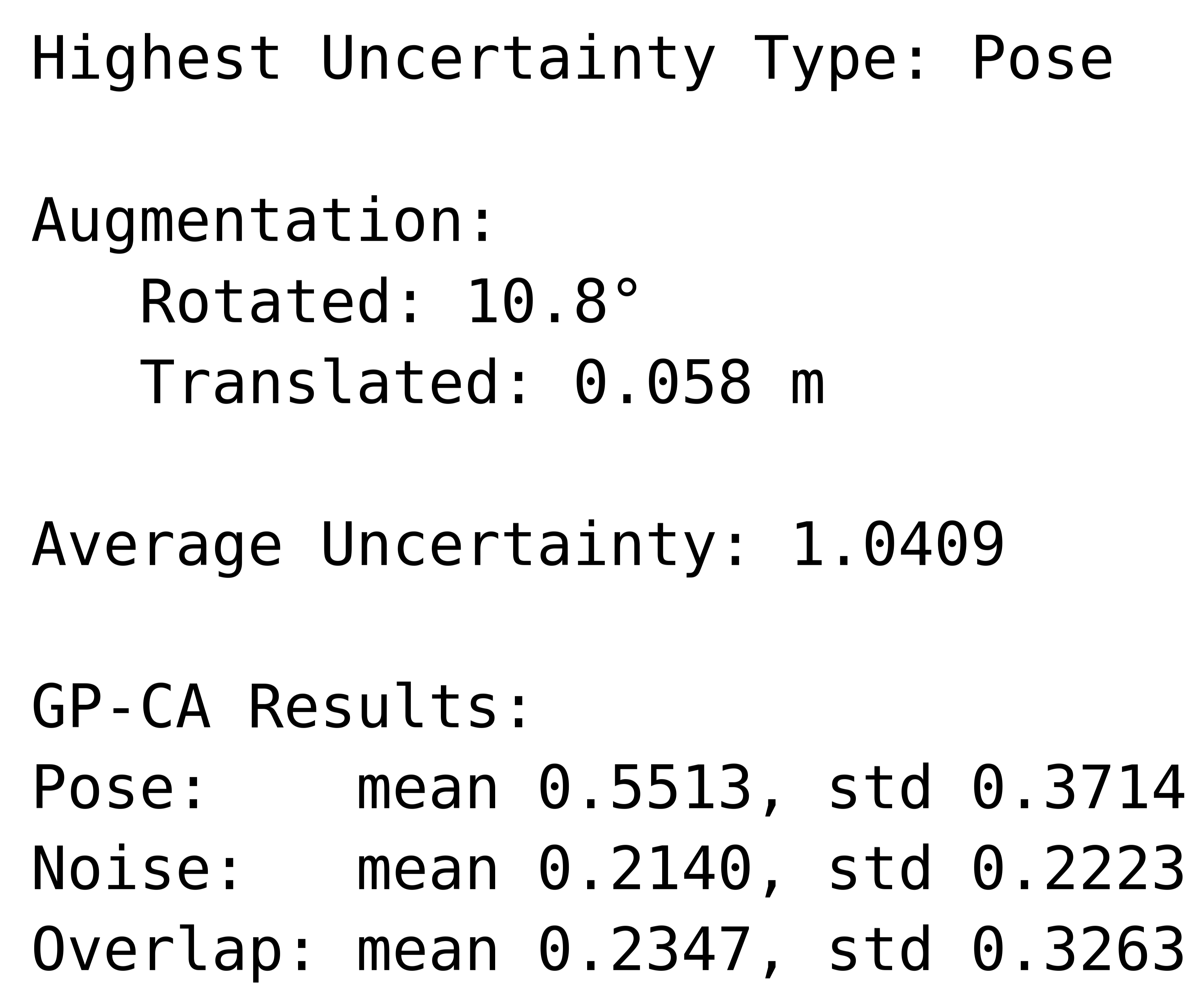}}%
  }
\end{tabular}
\caption{Real-robot validation example: (a) Input and ROI selection, (b) initial pose estimate, (c) ICP alignment with axes, (d) concept attribution with ground truth augmentation and GP-CA concept scores.}
\label{fig:robot}
\end{figure}

\begin{table}[t]
\centering
\caption{Per-component runtime (mean over 100 runs).}
\label{tab:runtime}
\small
\setlength{\tabcolsep}{3pt}
\renewcommand{\arraystretch}{0.95}
\begin{tabular}{lcc}
\toprule
\textbf{Component} & \textbf{Time} & \textbf{Device} \\
\midrule
ICP alignment & 3.0 s & CPU \\
DGCNN feature extraction & 1.2 s & GPU \\
GPC posterior sampling (1000 draws) & 10 ms & CPU \\
GP-CA score computation & 15 ms & CPU \\
BALD selection (optional) & 7 ms & CPU \\
GPC retraining (optional) & 10--30 s & CPU \\
\bottomrule
\end{tabular}
\end{table}

We verify the active learning results from Sec.~\ref{sec:rq2} on our experimental dataset. Because this set is noisier and more unstructured, the attainable accuracy ceiling is lower (85.2\% vs.~99\% on YCB). Evaluating label efficiency at 90\% would violate that ceiling and confound the comparison, we therefore report the labels needed to reach a target of 70\%. We evaluate BALD and random sampling under this criterion in five experiments, following the setup in Sec.~\ref{sec:rq2}. BALD requires 52 labels on average to reach 70\% accuracy versus 101 for random sampling; the relative gain is robust for nearby thresholds (65–75\%).
Fig.~\ref{fig:robot} illustrates the full loop on representative frames: from the input RGB-D image, we select a Region of Interest (ROI) (a) and deduce an initial 6-DoF pose estimate (b). Then, we perform the ICP alignment (c) while estimating the uncertainty and attributing it to causal concepts using GP-CA. The GP-CA panel (d) shows the attribution results.

We evaluate runtime on a workstation with an Intel i7-12700K CPU, 32\,GB RAM, and an NVIDIA RTX 3090 GPU. Tab.~\ref{tab:runtime} reports mean computation times over 100 runs. End to end runtime is dominated by ICP and DGCNN at about 4.2\,s per registration. Concept attribution and uncertainty scoring add roughly 25\,ms on the CPU, and BALD selection adds 7\,ms when enabled. GPC retraining takes 10\,s -- 30\,s and can run asynchronously, so the explainer overhead remains small while the geometric front end determines throughput. Real-world validation confirms GP-CA’s practical accuracy and efficiency in explaining ICP uncertainty.

\vspace{-5pt}
\begin{figure}[h]
  \centering
  \includegraphics[width=\columnwidth]{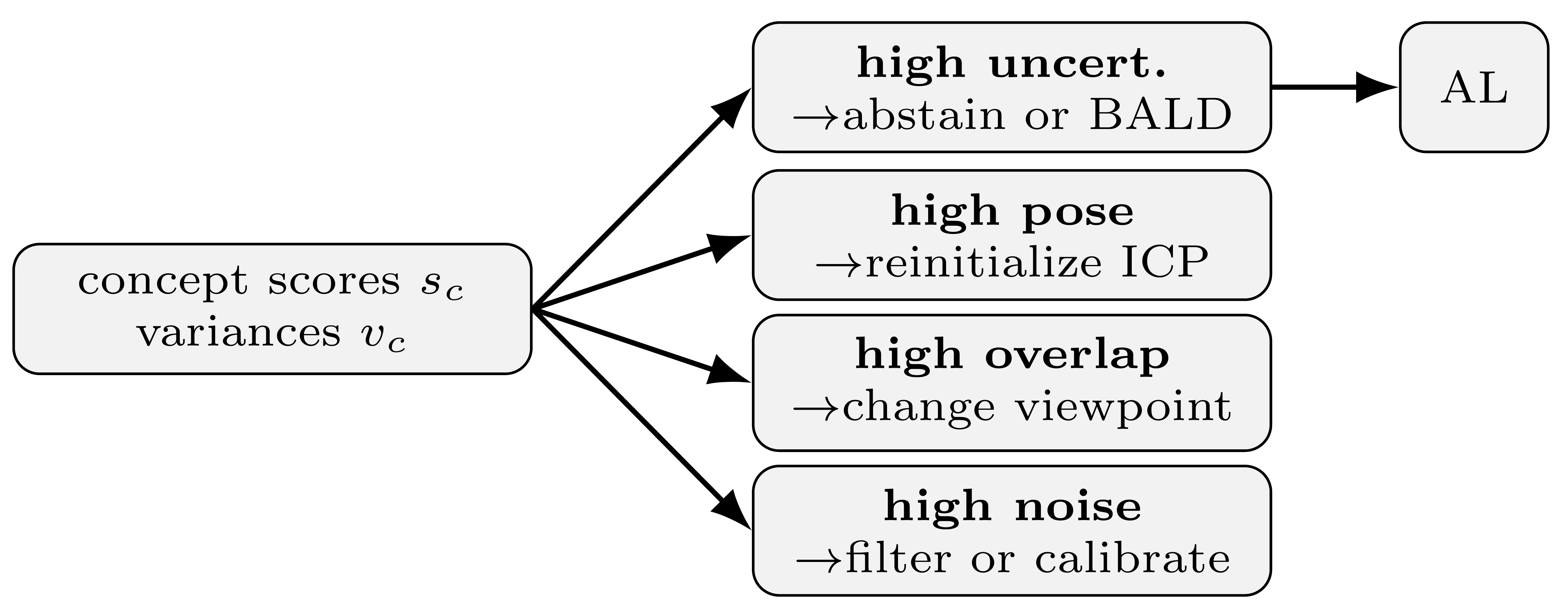}
  \caption{Mitigation mapping from concept outputs to actions. From the concept scores and variances, the system selects one of four branches: abstain/BALD, reinitialize ICP, change viewpoint, or filter/calibrate.}
  \label{fig:mitigation}
\end{figure}
\vspace{-5pt}

\section{Robotic Deployment for Action Recovery}

Fig.~\ref{fig:mitigation} summarizes possible actions that a robot can take, once it's able to explain uncertainty estimates. Given the concept score and its variance, we can trigger different actions. For example, in cases of high uncertainty, we can use active learning to retrain the GPC. A dominant pose score leads to reinitialization with a broader search. A dominant overlap score prompts a small viewpoint change, for example, via the active view planner from \cite{shi2025viso}, to increase common support. A dominant noise score entails stronger filtering and prompts a quick sensor check or calibration. After any action we recompute the scores and, if the improvement remains small for a few attempts, we queue the case for labeling so the explainer is updated. In the accompanying video, we illustrate some of these functionalities, which lead to robust robot behavior under uncertainty in point cloud registration.

\section{Conclusion}

We presented GP-CA, a novel concept-aware explanation approach for uncertainty in point cloud registration. We use a DGCNN to extract feature embeddings from aligned point clouds, which are passed to a GPC to produce probabilities over human-interpretable uncertainty concepts. GP-CA is also equipped with active learning to incorporate new concepts. We validate GP-CA on three publicly available datasets and in real-world experiments. Our results show that it quantifies uncertainty, attributes it to specific concepts, and produces estimates that align with distributional shift, enabling active learning with few queries and data points.

While effective, GP-CA assumes a predefined concept set and single-label attribution per instance (favoring a single dominant concept rather than explicit multi-cause labels), and its uncertainty captures posterior (epistemic) rather than aleatoric effects, some causes may be ambiguous from a single frame (e.g., pose vs.\ overlap). Performance depends on the embedding and on upstream ICP outputs, and training cost grows with labeled data and inducing inputs, though periodic recalibration and sparse updates mitigate these factors.

Overall, GP-CA outperforms representative baselines and facilitates more robust completion of given perception tasks by bringing explainable AI to geometric computer vision.

\section{Acknowledgement}
The authors thank Zicheng Guo from the Karlsruhe Institute of Technology for their valuable contribution to this work, in particular for the collection of real-world training data.

\bibliographystyle{IEEEtran}
\bibliography{thesis}

\begin{thebibliography}{10}
\providecommand{\url}[1]{#1}
\csname url@samestyle\endcsname
\providecommand{\newblock}{\relax}
\providecommand{\bibinfo}[2]{#2}
\providecommand{\BIBentrySTDinterwordspacing}{\spaceskip=0pt\relax}
\providecommand{\BIBentryALTinterwordstretchfactor}{4}
\providecommand{\BIBentryALTinterwordspacing}{\spaceskip=\fontdimen2\font plus
\BIBentryALTinterwordstretchfactor\fontdimen3\font minus \fontdimen4\font\relax}
\providecommand{\BIBforeignlanguage}[2]{{%
\expandafter\ifx\csname l@#1\endcsname\relax
\typeout{** WARNING: IEEEtran.bst: No hyphenation pattern has been}%
\typeout{** loaded for the language `#1'. Using the pattern for}%
\typeout{** the default language instead.}%
\else
\language=\csname l@#1\endcsname
\fi
#2}}
\providecommand{\BIBdecl}{\relax}
\BIBdecl

\bibitem{Besl1992}
P.~Besl and N.~McKay, ``A method for registration of 3-d shapes,'' \emph{IEEE PAMI}, vol.~14, no.~2, pp. 239--256, 1992.

\bibitem{zhang2014loam}
J.~Zhang, S.~Singh \emph{et~al.}, ``Loam: Lidar odometry and mapping in real-time.'' in \emph{RSS}, Berkeley, CA, USA, July 2014, pp. 1--9.

\bibitem{izadi2011kinectfusion}
S.~Izadi \emph{et~al.}, ``Kinectfusion: Real-time 3d reconstruction and interaction using a moving depth camera,'' in \emph{UIST}, Santa Barbara, CA, USA, October 2011, pp. 559--568.

\bibitem{sundermeyer2018implicit}
M.~Sundermeyer \emph{et~al.}, ``Implicit 3d orientation learning for 6d object detection from rgb images,'' in \emph{ECCV}, Munich, Germany, September 2018, pp. 699--715.

\bibitem{rusinkiewicz2001}
S.~Rusinkiewicz and M.~Levoy, ``Efficient variants of the icp algorithm,'' in \emph{3DIM}, Quebec City, Canada, June 2001, pp. 145--152.

\bibitem{Maken2021}
F.~Maken, F.~Ramos, and L.~Ott, ``Stein icp for uncertainty estimation in point cloud matching,'' \emph{RA-L}, vol.~7, no.~2, pp. 1063--1070, 2022.

\bibitem{yang2020teaser}
H.~Yang, J.~Shi, and L.~Carlone, ``Teaser: Fast and certifiable point cloud registration,'' \emph{TRO}, vol.~37, no.~2, pp. 314--333, 2021.

\bibitem{de_maio_deep_2022}
A.~De~Maio and S.~Lacroix, ``Deep bayesian icp covariance estimation,'' in \emph{ICRA}, Philadelphia, PA, USA, May 2022, pp. 6519--6525.

\bibitem{Landry2019}
D.~Landry, F.~Pomerleau, and P.~Gigu{\`e}re, ``Cello-3d: Estimating the covariance of icp in the real world,'' in \emph{ICRA}, Montreal, Canada, May 2019, pp. 8190--8196.

\bibitem{Bonnabel2016}
S.~Bonnabel, M.~Barczyk, and F.~Goulette, ``On the covariance of icp-based scan-matching techniques,'' in \emph{ACC}, Boston, MA, USA, July 2016, pp. 5498--5503.

\bibitem{Censi2007}
A.~Censi, ``An accurate closed-form estimate of {ICP}'s covariance,'' in \emph{ICRA}, Rome, Italy, April 2007, pp. 3167--3172.

\bibitem{Qin2024}
Z.~Qin, J.~Lee, and R.~Triebel, ``Towards explaining uncertainty estimates in point cloud registration,'' 2024, arXiv:2412.20612.

\bibitem{narr2016stream}
A.~Narr, R.~Triebel, and D.~Cremers, ``Stream-based active learning for efficient and adaptive classification of 3d objects,'' in \emph{ICRA}, Stockholm, Sweden, May 2016, pp. 227--233.

\bibitem{lee2025clever}
J.~Lee \emph{et~al.}, ``Clever: Stream-based active learning for robust semantic perception from human instructions,'' \emph{IEEE RA-L}, vol.~10, no.~9, pp. 8906--8913, 2025.

\bibitem{Hinterstoisser2012}
S.~Hinterstoisser \emph{et~al.}, ``Model based training, detection and pose estimation of texture-less 3d objects in heavily cluttered scenes,'' in \emph{ACCV}, Daejeon, Republic of Korea, November 2012, pp. 548--562.

\bibitem{Xiang2018}
Y.~Xiang \emph{et~al.}, ``Posecnn: A convolutional neural network for 6d object pose estimation in cluttered scenes,'' in \emph{RSS}, Pittsburgh, PA, USA, June 2018.

\bibitem{Lai2011}
K.~Lai \emph{et~al.}, ``A large-scale hierarchical multi-view {RGB-D} object dataset,'' in \emph{ICRA}, Shanghai, China, May 2011, pp. 1817--1824.

\bibitem{Chen1992}
Y.~Chen and G.~Medioni, ``Object modelling by registration of multiple range images,'' \emph{Image Vis. Comput.}, vol.~10, no.~3, pp. 145--155, 1992.

\bibitem{segal_generalized-icp_2009}
A.~Segal, D.~Haehnel, and S.~Thrun, ``Generalized-icp,'' in \emph{RSS}, Seattle, WA, USA, July 2009.

\bibitem{ali2023explainable}
S.~Ali \emph{et~al.}, ``Explainable artificial intelligence (xai): What we know and what is left to attain trustworthy artificial intelligence,'' \emph{Inf. Fusion}, vol.~99, p. 101805, 2023.

\bibitem{lundberg_unified_2017}
M.~Lundberg and S.-I. Lee, ``A unified approach to interpreting model predictions,'' in \emph{NeurIPS}, Long Beach, CA, USA, December 2017, pp. 4765--4774.

\bibitem{marino_methodology_2008}
S.~Marino \emph{et~al.}, ``A methodology for performing global uncertainty and sensitivity analysis in systems biology,'' \emph{J. Theor. Biol.}, vol. 254, no.~1, pp. 178--196, 2008.

\bibitem{Kim2018}
B.~Kim \emph{et~al.}, ``Interpretability beyond feature attribution: Quantitative testing with concept activation vectors (tcav),'' in \emph{ICML}, Stockholm, Sweden, July 2018, pp. 2668--2677.

\bibitem{Houlsby2011}
N.~Houlsby \emph{et~al.}, ``Bayesian active learning for classification and preference learning,'' 2011, arXiv:1112.5745.

\bibitem{Wang2019}
Y.~Wang \emph{et~al.}, ``Dynamic graph cnn for learning on point clouds,'' \emph{ACM Trans. Graph.}, vol.~38, no.~5, p. 146, 2019.

\bibitem{Qi2017}
C.~R. Qi \emph{et~al.}, ``Pointnet: Deep learning on point sets for 3d classification and segmentation,'' in \emph{CVPR}, Honolulu, HI, USA, July 2017, pp. 652--660.

\bibitem{Qi2017b}
C.~R. Qi \emph{et~al.}, ``Pointnet++: Deep hierarchical feature learning on point sets in a metric space,'' in \emph{NeurIPS}, vol.~30, Long Beach, CA, USA, December 2017, pp. 5099--5108.

\bibitem{Rasmussen2006}
C.~Rasmussen and C.~Williams, \emph{Gaussian Processes for Machine Learning}.\hskip 1em plus 0.5em minus 0.4em\relax Cambridge, MA, USA: MIT Press, 2006.

\bibitem{hensman2015}
J.~Hensman, A.~Matthews, and Z.~Ghahramani, ``Scalable variational gaussian process classification,'' in \emph{AISTATS}, San Diego, CA, USA, May 2015, pp. 351--360.

\bibitem{Grimmett2016}
H.~Grimmett \emph{et~al.}, ``Introspective classification for robot perception,'' \emph{IJRR}, vol.~35, no.~7, pp. 743--762, 2016.

\bibitem{scikit-learn}
F.~Pedregosa \emph{et~al.}, ``Scikit-learn: Machine learning in {P}ython,'' \emph{Journal of Machine Learning Research}, vol.~12, pp. 2825--2830, 2011.

\bibitem{Qian2022}
G.~Qian \emph{et~al.}, ``Pointnext: Revisiting {PointNet++} with improved training and scaling strategies,'' in \emph{NeurIPS}, New Orleans, LA, USA, December 2022, pp. 23\,192--23\,204.

\bibitem{Jolliffe2002}
I.~T. Jolliffe, \emph{Principal Component Analysis}, ser. Springer Series in Statistics.\hskip 1em plus 0.5em minus 0.4em\relax New York, NY, USA: Springer, 2002.

\bibitem{mcinnes_umap_2018}
L.~McInnes, J.~Healy, and J.~Melville, ``Umap: Uniform manifold approximation and projection for dimension reduction,'' \emph{J. Open Source Softw.}, vol.~3, no.~29, p. 861, 2018.

\bibitem{Hinton2006}
G.~Hinton and R.~Salakhutdinov, ``Reducing the dimensionality of data with neural networks,'' \emph{Science}, vol. 313, no. 5786, pp. 504--507, 2006.

\bibitem{Wen2024}
B.~Wen \emph{et~al.}, ``Foundationpose: Unified 6d pose estimation and tracking of novel objects,'' in \emph{CVPR}, Seattle, WA, USA, June 2024, pp. 17\,868--17\,879.

\bibitem{shi2025viso}
Y.~Shi \emph{et~al.}, ``Viso-grasp: Vision-language informed spatial object-centric 6-dof active view planning and grasping in clutter and invisibility,'' 2025, arXiv:2503.12609.

\end{thebibliography}

\end{document}